\pdfoutput=1
\documentclass[11pt]{article}
\usepackage[]{acl2021}

\usepackage{times}
\usepackage{latexsym}

\usepackage[T1]{fontenc}
\usepackage[utf8]{inputenc}
\usepackage{microtype}

\usepackage{calc}
\usepackage{graphicx}
\usepackage{tabularx}
\usepackage{multirow}
\usepackage{booktabs}
\usepackage{latexsym}
\usepackage{pifont}
\usepackage{algorithm}
\usepackage{algpseudocode}
\usepackage{bbm}
\usepackage{amsthm}
\usepackage{amsmath}
\usepackage{amssymb}
\usepackage{ulem}
\DeclareMathOperator*{\argmax}{argmax}

\usepackage{microtype}

\newcolumntype{L}[1]{>{\raggedright\let\newline\\\arraybackslash\hspace{0pt}}m{#1}}
\newcolumntype{C}[1]{>{\centering\let\newline\\\arraybackslash\hspace{0pt}}m{#1}}
\newcolumntype{R}[1]{>{\raggedleft\let\newline\\\arraybackslash\hspace{0pt}}m{#1}}

\definecolor{navyblue}{rgb}{0.0, 0.0, 0.5}
\definecolor{cadmiumorange}{rgb}{0.93, 0.53, 0.18}
\definecolor{darkspringgreen}{rgb}{0.09, 0.45, 0.27}
\definecolor{chamoisee}{rgb}{0.63, 0.47, 0.35}
\definecolor{ochre}{rgb}{0.8, 0.47, 0.13}
\definecolor{greenrgb}{rgb}{0.18, 0.71, 0.18}
\title{How ``Multi'' is Multi-Document Summarization?}

\author{Ruben Wolhandler\thanks{\; Equal contribution.} \quad 
        Arie Cattan\footnotemark[1] \quad
        Ori Ernst \quad 
        Ido Dagan \\ 
        Computer Science Department, Bar Ilan University \\ 
 {\normalsize\tt  \{rwolhandler,arie.cattan,oriern\}@gmail.com} \quad \normalsize\tt dagan@cs.biu.ac.il \\
 }

\date{}

\begin{document}
\maketitle

\begin{abstract}
The task of multi-document summarization (MDS) aims at models that, given multiple documents as input, are able to generate a summary that combines disperse information, originally spread \textit{across} these documents. Accordingly, it is expected that both reference summaries in MDS datasets, as well as system summaries, would indeed be based on such dispersed information. In this paper, we argue for quantifying and assessing this expectation. To that end, we propose an automated measure for evaluating the degree to which a summary is ``disperse'', in the sense of the number of source documents needed to cover its content. We apply our measure to empirically analyze several popular MDS datasets, with respect to their reference summaries, as well as the output of state-of-the-art systems. Our results show that certain MDS datasets barely require combining information from multiple documents, where a single document often covers the full summary content. Overall, we advocate using our metric for assessing and improving the degree to which summarization datasets require combining multi-document information, and similarly how summarization models actually meet this challenge.\footnote{Our code is available in \url{https://github.com/ariecattan/multi_mds}.}

\end{abstract}

\section{Introduction}

% \begin{table}[]
%     \centering
%     \resizebox{0.48\textwidth}{!}{
%     \begin{tabular}{p{0.6\textwidth}}
%     \toprule

%     \textbf{Doc 1}: The Pompidou Centre in Paris hopes to display a long-vanished Picasso painting in May, now that it has been recovered by U.S. customs authorities. This undated photo provided by the United States Department of Justice, shows a cubist painting entitled “The Hairdresser” by Pablo Picasso... \\ 
    
%     \midrule 
    
%     \textbf{Doc 2}: A stolen Picasso worth millions of dollars was shipped to the U.S. in a package that described it as a \$37 "art craft" — but it will soon be on its way back to France. Federal prosecutors in Brooklyn filed papers Thursday to forfeit the century-old cubist painting, which was swiped from a museum storeroom in 2001.  \\
    
%     \midrule
    
%     \textbf{Doc 3}: A Picasso painting missing from Paris for more than a decade resurfaced in the United States, where it had been shipped under false pretenses as a \$37 holiday... \\
    
%     \midrule
    
% %     \textbf{Summary}: \textbf{\textcolor{navyblue}{A Picasso painting that was found to have vanished from a Paris museum has turned-up in the US}}, \textbf{\textcolor{ochre}{in a package shipped from Belgium.}} \\

% %          \bottomrule 
         
% %     \end{tabular}}
% %     \caption{\ariec{PLACEHOLDER for a better example} An example of a summary of multiple documents. The fact XX and XX appear in all source documents, while XX and XX are mentioned only in Doc XX and XX respectively.}
% %     \label{tab:example}
% % \end{table}

\begin{table}[]
    \centering
    \resizebox{0.48\textwidth}{!}{
    \begin{tabular}{p{0.6\textwidth}}
    \toprule

    \textbf{Doc 1}: \textcolor{ochre}{Indigenous Arctic people} urged European countries to step up the fight against global warming, \textcolor{ochre}{saying it is threatening their societies}.
    The Arctic Council said that \textcolor{navyblue}{the amount of sea ice around the North Pole has decreased about 8 percent in 30 years because of global warming}\\ 
    
    \midrule 
    
    \textbf{Doc 2}: One of the topics discussed at the global warming conference is \textcolor{navyblue}{the decrease of the sea ice in the Arctic}.. \\
    
    \midrule
    
    \textbf{Doc 3}: Glaciologists worry most about \textcolor{navyblue}{the Arctic ice sheet: if gradually melted}, \textcolor{red}{to raise ocean levels worldwide by about five meters stems directly } \textcolor{navyblue}{from global warming or from more localized conditions}. \\
    
    \midrule
    
    \textbf{Summary}: \textcolor{navyblue}{Global warming has caused the Arctic ice to melt considerably.} \textcolor{ochre}{These changes are threatening the indigenous Arctic population} \textcolor{red}{and could raise ocean levels worldwide.}  \\

         \bottomrule 
         
    \end{tabular}}
    \caption{ An example of a summary of multiple documents. The proposition ``melting ice'' (in blue) appears in all source documents, while ``the threat for the Arctic population'' (in ochre) and ``the rising water'' (in red) are mentioned only in documents 1 and 3 respectively.}
    \label{tab:example}
\end{table}

Multi-document Summarization (MDS) consists of creating a short and concise summary that includes the salient information in a set of related documents. Beyond the challenges in single-document summarization, a summary of multiple texts is expected to 
combine and assemble information spread \textit{across} several input texts. Table~\ref{tab:example} illustrates such an example where the summary 
combines multiple facts from the different documents about global warming. While the main fact (``melting ice'') can be described in all source documents, secondary information such as ``the rising water'' often appear only in certain document(s). 

In order to develop MDS models that effectively merge information from various sources, it is necessary that reference summaries in MDS datasets should be based on such dispersed information across the source documents. However, to the best of our knowledge, while existing datasets assume that this property is realized, measuring (automatically) the degree of multi-text merging was not investigated in the literature. 

In this work, we suggest quantifying the degree to which a summary is ``disperse'' in terms of the minimum number of documents needed to cover its content. Accordingly, we develop an automated method for measuring this aspect for any MDS summary. To that end, we first identify the potential provenance of the summary information in all source documents. Then, for each possible number of documents, we form the subset of documents that includes the largest amount aligned information with the summary. Finally, we define the degree of multi-text merging of an MDS summary as a function of the amount of summary information \emph{not} covered by each subset of documents.

We apply our automated measure to evaluate the degree of multi-text merging in four prominent MDS datasets (DUC, TAC, MultiNews and WCEP) as well as the output of five recent systems. Our results show that some existing datasets barely involve multi-text merging because the reference summary information mostly appears in a single document. Unsurprisingly, the length of the summary has a substantial impact on the amount of multi-text merging since longer summaries cover more detailed information which tends to be spread across documents.

Taken together, our work is the first to measure and empirically analyze multi-text merging in MDS datasets and model summaries. We suggest that future work will use our methodology to develop better datasets and to improve the degree of multi-text merging in MDS models.

\section{A Measure for Multi-text Merging}
\label{sec:metric}

\subsection{Motivating Analysis}
\label{subsec:motivation}

The common dataset structure for an MDS instance is a topic that consists of a set of source documents $D = \{D_1, ..., D_n\}$ and a summary $S$. To motivate our measure, we first analyze the degree of multi-text merging on a sample of topics. To that end, we leverage the  Summary-Source-Alignment dataset of \citet{ernst-etal-2021-summary}, in which human annotators aligned all propositions in reference summaries with corresponding propositions in the source documents that cover the same information, as exemplified in Table~\ref{tab:example}. Given these alignments on 9 MDS topics from MultiNews~\citep{fabbri-etal-2019-multi}, each composed of 4 source documents, we find that a single source document suffices to cover alone 70\% of the summary propositions while 2 documents cover 95\% of them. The remaining source documents thus hardly provide any substantial information to the summary.

Motivated by this analysis, we develop an automated measure that allows to evaluate the degree of multi-text merging in entire MDS datasets and in systems summaries. 
% Likewise, we suggest to apply our measure on system MDS summaries in order to measure how different systems manage to merge information across different documents. 
Our measure operates in the following steps. We first define the coverage score for a given subset of source documents~(§\ref{subsec:contribution}). Then, to approximate the minimum number of documents required to cover increasing portions of the summary information, we greedily construct, for each possible number of source documents, the subset of source documents with the highest coverage score~(§\ref{subsec:greedy}). Finally, we measure the total amount of summary information in all subset sizes, yielding a corresponding coverage curve~(§\ref{subsec:aac}).

\subsection{Relative Coverage Score}
\label{subsec:contribution}

Let $D^\ast$ be a subset of source documents $D^\ast \subseteq D$. We define the \emph{relative coverage} of $D^\ast$ as the proportion of information that is covered by $D^\ast$, normalized by the information covered by all source documents $D$:

\begin{equation}
    cov(D^\ast, D, S) = \frac{s(D^\ast, S)}{s(D, S)} 
    \label{eq:cov}
\end{equation}

% where $cov(D^\ast, S)$ is an affinity measure that quantifies how much information of the summary $S$ is included in the subset $D^\ast$.
% We consider two variants of $s(D^\ast, S)$ ($D$ is by definition also a subset $D^\ast$, which contains all source documents): \superpal{} and \oracle{}.
% If all the summary information is included in one document $D_i$, the contribution of the subset $D^\ast = \{D_i\}$ is 1. 

For the absolute coverage score $s(D^\ast, S)$, we aim to approximate the human annotation of summary-source proposition alignment in~\citep{ernst-etal-2021-summary}, which is based on the well established Pyramid scheme~\citep{nenkova-passonneau-2004-evaluating}. Specifically, we follow their automated scheme: (1) we extract all propositions from the summary and all source documents using OpenIE~\citep{Banko2008OpenIE},\footnote{We use the AllenNLP implementation of~\citep{stanovsky-etal-2018-supervised} to extract the OpenIE tuples. Following~\citet{ernst-etal-2021-summary, Ernst2021APC}, we convert each OpenIE tuple into a proposition string by concatenating the predicate and its arguments by their original order.} (2) we compute the similarity score between the propositions in the summary and the source documents using \textsc{SuperPAL}, an NLI model fine-tuned on proposition alignment~\citep{ernst-etal-2021-summary}, (3) $s(D^\ast, S)$ is defined as the number of propositions in $S$ that are aligned with some proposition in $D^\ast$.

We consider the proportion $s(D^\ast, S){}/{s(D, S)}$ and not the absolute coverage $s(D^\ast, S)$ for two main reasons. First, as both reference and system summaries are known to include hallucinated information~\citep{maynez-etal-2020-faithfulness}, we need to discard them in our measure in order to properly estimate the amount of information that each single source document actually provides to the summary. Second, normalizing the coverage score will mitigate the potential omissions of the alignment model.

\subsection{Maximally-Covering Document Subsets}
\label{subsec:greedy}

Given an MDS topic with $n$ source documents, we aim to measure the maximal content coverage of the summary content by a document subset of size $k\leqslant n$.
To that end, we form $n$ subsets of source documents $\{D^\ast_1, ..., D^\ast_n\}$, such that each subset $D^\ast_k$ includes an optimized set of $k$ source documents, covering the maximal amount of summary information. Specifically, we employ a greedy approach where we add one source document at a time to maximize the increase in the relative coverage score (Eq.~\ref{eq:cov}), as follows:\footnote{We also tested the (exponential) optimal approach for finding $D^\ast_k$ on two datasets where this was feasible, namely Multi-news and DUC 2003-2004, and found no significant difference in AAC scores vs. the greedy approach (up to 0.1 difference), suggesting its sufficiency.},

\begin{equation}
        D^\ast_{k+1} =  \argmax_{d \in D \setminus D^\ast_{k}}{cov(D^\ast_k \cup d, D, S)}     
\end{equation}

 That is, $D^\ast_1$ includes the single document $d_1$ with the maximal coverage score, $D^\ast_2$ adds the document $d$ that marginally contributes the most to the coverage, and so on.

\subsection{Dispersion Score: Area Above the Curve}
\label{subsec:aac}

Given the $n$ optimal subsets (in the above greedy sense) of source documents for a given topic $t$, let $cov^t_k$ be the coverage of the subset $D^\ast_k$. We then define the overall coverage for an entire MDS dataset as the average coverage score for each topic: $cov_k = \frac{1}{T}\sum_{t \in T}{cov^t_k}$. Accordingly, $cov_k$ aims to measure the average amount of summary information that $k$ source documents cover. 

To allow qualitative analysis of the degree of multi-text merging, we plot the curve of $cov_k$ for $k \in \left[1, n\right]$, as illustrated in Figure~\ref{fig:superpal}. Since the optimal subsets are formed incrementally, $cov_k$ is a monotonic non-decreasing function whose maximum is $1$. As shown in Figure~\ref{fig:superpal}, the curve corresponding to WCEP-10 is close to $1$ already when $k=1$ (single source document), while the curve of DUC 2006-2007 gradually increases to 1, indicating that a larger number of source documents are required to cover the summary content.

While visual plots are insightful, we are also interested in \emph{quantifying} how slowly $cov_k$ converges to 1. Analogously to $cov_k$, the difference $1-cov_k$, expressed by the area above the $cov_k$ curve, aims to measure the average amount of summary information that $k$ documents do \emph{not} cover. Therefore, we define our ``dispersion'' score as the Area Above the $cov_k$ Curve (AAC).\footnote{It is easy to see that this definition of the AAC score is equivalent to the average of the individual per-topic AAC scores across the dataset.} The higher the AAC, the more multi-text merging is required. To properly compare the degree of multi-text merging on various datasets with different numbers of source documents, we normalize the AAC by the maximum number of required source document, $n_{\textsc{max}}$.\footnote{In particular, we set $n_{\textsc{max}}=10$ based on Figure~\ref{fig:superpal}.} The exact formula of the AAC becomes:\footnote{Technically, our formula for calculating the AAC score considers the area above the \emph{non-interpolated} curve. This is consistent with the common view by which the Average Precision score is often considered as the area under the non-interpolated recall-precision curve.}

\begin{equation}
AAC = \frac{1}{n_{\textsc{max}}} \sum_{k=1}^{n}{1 - cov_k} 
\end{equation}

It is important to note that we avoid normalizing our dispersion metric  by the number of source documents in a topic, aiming to better reflect the absolute degree of dispersity. To illustrate, consider one dataset with say 3 documents per cluster, where 2 are sufficient for covering the reference, vs. a dataset with 20 documents per cluster, where 5 are sufficient. Our (non-normalized) score would clearly favor the second dataset, fitting our main motivation in this paper, while a normalized score would counter-productively favor the first.

\subsection{Related Evaluation Practices}
\label{subsec:related}

When evaluating a new summarization model, beyond lexical similarity between the system and reference summary (ie. ROUGE), it is common to also report specific aspects such as relevance and informativeness~\citep{jung-etal-2019-earlier, peyrard-2019-simple}, faithfulness~\citep{kryscinski-etal-2020-evaluating, scialom-etal-2021-questeval}, grammar, redundancy~\citep{Lloret2013TacklingRI, xiao-carenini-2020-systematically}, or sentence fusion~\citep{lebanoff-etal-2019-analyzing}. Other aspects such as extractiveness or compression were investigated for reference summaries~\citep{grusky-etal-2018-newsroom}.

Our dispersion measure aims to measure, both for reference and system summaries, the degree of multi-text merging, which is an essential aspect in multi-document summarization.

\section{Empirical Analysis}
\label{subsec:emperical}

\begin{figure}[t]
    \centering
    \includegraphics[width=.48\textwidth]{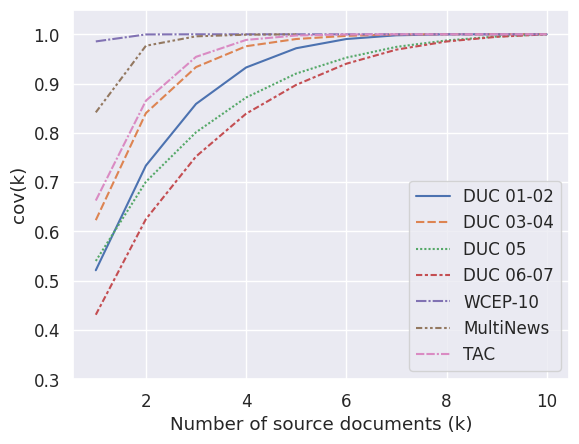}
    \caption{Curves of $cov_k$ on MDS datasets. Slow convergence to 1, as for DUC 06-07, reflects highe degree of multi-text merging.
    }
    \label{fig:superpal}
\end{figure}

To assess our automated dispersion measure, we apply our method on the exact same MultiNews topics that we used for our motivating analysis~(§\ref{subsec:motivation}). As illustrated in Appendix~\ref{app:human_comparison}, we observe a similar behaviour in our automated measure, assessing its effectiveness. We cannot evaluate correlation due to a lack of statistical significance, as we have annotations only on 9 topics.

\begin{table}[t]
    \centering
    \resizebox{0.48\textwidth}{!}{
    \begin{tabular}{llccc}
    \toprule
    % data - blank - r1 - r1 - blank - r2 - r2- blank - s - s
    && \#docs &  length & AAC \\
    \midrule 
    WCEP-10         && 10   & 28        & 0.1 (±0.7)     \\
    MultiNews       && 2.8  & 218       & 1.8 (±2.3)    \\ 
    TAC             && 10   & 100       & 5.4 (±4.1)     \\ 
    DUC 2003-04     && 10   & 102       & 6.6 (±5.0)    \\ 
    DUC 2001-02     && 10   & 235       & 10.2 (±5.7)     \\
    DUC 2005        && 30   & 250       & 13.4 (±9.6)    \\ 
    DUC 2006-07     && 25   & 250       & 16.5 (±9.1)    \\
    \bottomrule

    \end{tabular}}
\caption{AAC scores and standard deviation of the reference summary in several datasets. We group the DUCs with equal summary's length (number of tokens) and with equal number of source documents.} 
\label{tab:dataset_results}
\end{table}

\paragraph{Reference Summaries}

We compute the AAC scores on the reference summaries of DUC 2001 to 2007 \citep{nist2014duc}, TAC 2008 to 2011 \citep{nisttac}, MultiNews~\citep{fabbri-etal-2019-multi} and WCEP-10 \citep{gholipour-ghalandari-etal-2020-large}. We report the AAC score for each dataset, as well as the standard deviation of the individual AAC scores across the dataset topics, in Table~\ref{tab:dataset_results} and plot the $cov_k$ curves in Figure~\ref{fig:superpal}.

The AAC score is 0.1 for WCEP-10 and 1.8 for MultiNews, showing a rather poor degree of multi-text merging, where the summary information mostly appears in a single document~(see Figure~\ref{fig:superpal}). Early datasets such as DUC and TAC exhibit a higher degree of multi-text merging, with the highest 16.5 AAC score for DUC 06-07. We also observe, for all datasets, that the standard deviations of the per-topic AAC scores are of substantial magnitudes, proportional to the average AAC score across the dataset.
This indicates that the various topics in each dataset present varying levels of dispersity, as can be expected in real-world scenarios.

% revealing a poor degree of multi-text merging. In fact, all the summary information usually appears in a single document, as we can see it in Figure~\ref{fig:superpal} with a $cov_1$ score close to 1.

% MultiNews attains the second lowest AAC score: 1.8, showing a rather low degree of multi-text merging. In fact, as shown in Figure~\ref{fig:superpal}, $cov_1$ is approximately 0.84, which demonstrates that a single document covers most of the summary information.

% Reference summaries in early datasets such as DUC (2001 to 2007) and TAC (2008 to 2011) are much more disperse and achieves higher AAC scores. Indeed, as we can see in Figure~\ref{fig:superpal}, DUC 2001 and 2002 require 6 documents until $cov$ reaches 1, meaning that at least 6 source documents are needed to cover all the information in the reference summary. Similarly DUC 2005, 2006 and 2007 require about 9 documents.

Further, we find that the length of the summary has an impact on the degree of multi-text merging. The longer the reference summary the more diverse it becomes. Indeed, DUC datasets with summary length of 100 tokens have a lower AAC score than those with length of more than 200 tokens. These results seem intuitive because the more concise the summary, the more salient the information in it is, and hence is more likely to appear in most source documents (e.g ``melting ice'' in Table~\ref{tab:example}). Nevertheless, the summary length is not the only factor, as on MultiNews we obtain a low AAC score, whereas the reference summary length is longer than 200 tokens on average.

\begin{table}[t]
    \centering
    \resizebox{0.4\textwidth}{!}{
    \begin{tabular}{llcclcc}
    \toprule

     && DUC 2004 & TAC 2011 \\
    \midrule 
    Reference            &&  \textbf{7.2} (±5.3)   & \textbf{4.6} (±3.7)   \\ 
    \midrule
    RL-MMR             && 2.6 (±2.4)   &  1.9 (±2.9) \\ 
    PG-MMR             && 2.5 (±2.6)   &  2.8 (±3.2) \\ 
    LexRank            && 3.7 (±3.3)   &  3.2 (±2.9) \\
    PG                 && 4.7 (±3.1)  &  3.1 (±2.6) \\ 
    ProCluster         && 4.1 (±3.7)   &  3.9 (±4.4) \\  
      
    %   Human             && 39.5 & 34.3 & 86.8 && 13.2 & 10.2 & 78.0  && 93.3 & 75.6 & 80.6  \\ 
      \bottomrule
    \end{tabular}}
\caption{AAC scores and standard deviation on different systems. For both DUC and TAC, reference summaries are more disperse than system summaries.}
\label{tab:systems}
\end{table}

\paragraph{System Summaries}
We also compute the AAC scores on the output of several system summaries: LexRank \citep{Erkan2004LexRankGL}, PG \citep{see-etal-2017-get}, PG-MMR \citep{lebanoff-etal-2018-adapting}, RL-MMR \citep{mao-etal-2020-multi} and ProCluster \citep{Ernst2021APC}, when tested on DUC 2004 and TAC 2011.
The AAC scores are presented in Table~\ref{tab:systems} and the curves of $cov_k$ are shown in Appendix~\ref{app:graphs}. We notice that, for both DUC 2004 and TAC 2011, the AAC score of the reference summary is slightly higher than the AAC score of all systems summaries. The curves of systems summaries~(Appendix~\ref{app:graphs}) have similar trends to the curves of the reference summaries, which leads us to believe that the more disperse the dataset will be, the more disperse the development of the systems will be.

% What we would like to point out is that even when the salient information in an input cluster is disperse, in most cases it can still be covered by a not-too-large subset of the given documents. Accordingly, it may be worthwhile to explore MDS modeling architectures that first identify such a salience-covering subset, and then feed this reduced set to the summarization system, with the aim of easing the summarization step (which, for example, may be impactful for deep end-to-end models, that may struggle with very large inputs). We will suggest this perspective as a possible direction for future research.

\begin{table*}[t]
    \centering
    \resizebox{0.8\textwidth}{!}{
    \begin{tabular}{lccclccclccc}
    \toprule
     & \multicolumn{3}{c}{$D_{\textsc{all}}$}   && \multicolumn{3}{c}{$D^*_1$} && \multicolumn{3}{c}{$D^*_2$} \\
            \cmidrule{2-4} \cmidrule{6-8} \cmidrule{10-12}
            & R-1 & R-2 & R-L           && R-1 & R-2 & R-L  && R-1 & R-2 & R-L \\
        \midrule
        DUC     &  36.4 &  8.7 & 19.7 && 31.3 & 7.0 & 17.6 && 34.3 & 8.2 & 18.8 \\
        TAC    &  36.6 &  10.0 & 20.7 && 39.3 & 14.4 & 23.9 && 37.7 & 11.0 & 21.2  \\
        % MultiNews   &  49.9 & 20.9 & 25.8 &&  \\ 
        MultiNews (ours) & 47.9 & 19.4 & 25.0 && 48.0 & 19.5 & 25.2 && 48.3 & 19.6 & 25.1 \\ 
        WCEP-10     &  45.8 & 25.0 & 37.6 && 46.4 & 24.8 & 37.3 && 46.7 & 25.2 & 37.8 \\
        \bottomrule
    \end{tabular}}
    \caption{Results of \textsc{Primera} when fine-tuning on all source documents ($D_{\textsc{all}}$), the source document with highest coverage ($D_1^*$) and the two source documents with highest coverage ($D_2^*$). Our results on MultiNews slightly differ from the original \textsc{Primera} paper~\citep{xiao-etal-2022-primera} because we fine-tune with a smaller number of optimization steps (See App.~\ref{app:mds_sds})}
    \label{tab:sds}
\end{table*}

\paragraph{How MDS models benefit from training on multiple documents?} To answer this question, we fine-tune PRIMERA~\citep{xiao-etal-2022-primera}, a state-of-the-art MDS model, to generate the summary given only a single document or two source documents as input, on multiple datasets. For this experiment, we select the source document(s) with the highest coverage score (§\ref{subsec:contribution}), for both training and inference. More implementation details are presented in Appendix~\ref{app:mds_sds}. We report the results for each dataset in Table~\ref{tab:sds}. For all datasets except for DUC, models trained on a subset of documents achieve comparable or higher ROUGE scores than models trained on the entire dataset. This finding hints that it may be worthwhile to explore MDS modeling architectures that first identify such a salience-covering subset, and then feed this reduced set to the summarization system, with the aim of easing the summarization step.

% \section{Discussion}

\paragraph{Challenge: Applicability to non-news domains}

As a preliminary attempt to examine the applicability of our dispersity measure implementation to non-news domains, we examined MS\^{}2~\citep{deyoung-etal-2021-ms}, an MDS dataset from the biomedical domain. However, we observe that the results are not reliable for this type of data due to insufficient quality of the proposition alignment step, for two main reasons. First, abbreviations of technical terms are very common in scientific papers, which presents a challenge to the SuperPAL aligner that we use. For example, consider the document proposition \emph{``The use of a Quadriceps tendon graft in primary ACL reconstruction leads to equal or better functional outcomes.''}. In this case, SuperPAL failed to predict alignment with the summary  proposition \emph{``\textbf{QT} autograft represent a feasible option for primary ACL reconstruction.''}, which uses the abbreviation \emph{``QT''}. However, when we replaced the abbreviation with the full term \textit{``Quadriceps Tendon"}, which appears in the document proposition, then SuperPAL successfully predicted alignment. The abundance of such cases indicates the need to adopt proposition alignment tools to their target domain, particularly with respect to technical terminology and abbreviations.

A second challenge stems from our use of a proposition alignment tool in order to track the evidence supporting a certain summary proposition. In fact, taking this approach assumes that the evidence in the source document should entail the summary proposition, which is mostly the case in the news domain. However, we observed that in the scientific domain of our data, a summary proposition often synthesizes information from multiple source evidences, yielding a novel consolidating proposition that is entailed only from the aggregation of multiple source evidences. For example, a typical such situation occurs when source documents include contradictory evidences or different perspectives, which is typical in scientific and other types of texts, while the summary synthesizes this evidences by pointing at the disagreements. To address this type of cases, more complex mechanisms of evidence tracking would be needed in order to compute dispersity, rather than just using 1:1 proposition alignment for this purpose.

\section{Conclusion}
We propose a measure to evaluate the degree of multi-text merging, an essential aspect in multi-document summarization. Using this measure, we found that some prominent MDS datasets contain summaries that hardly combine information from multiple input sources. Furthermore, we show that fine-tuning on a single effective document, achieves better summary performance than fine-tuning on the full set of documents.

\section{Limitations}
As described in Section~\ref{subsec:aac}, we use the SuperPAL model~\citep{ernst-etal-2021-summary} to predict summary-source alignment scores. Similarly to recent model-based evaluation metrics (e.g FactCC, BERTScore, etc.), our measure can also include noise due to some inaccurate predictions or proposition extraction. In addition, the SuperPAL model is time- and computation hardware- consuming because it assigns a BERT-based score for every summary-source proposition pairs.

\section*{Acknowledgments}
We thank the anonymous reviewers for their insightful comments. This work was supported in part by Intel Labs, the Israel Science Foundation grant 2827/21 and by a grant from the Israel Ministry of Science and Technology. Arie Cattan is partially supported by a fellowship for excellence in data science from the Bar-Ilan Data Science Institute (funded by the Israeli PBC).

\bibliographystyle{acl_natbib}
\bibliography{anthology,acl2021}

\clearpage
\appendix

\section{Datasets}

Here, we describe the datasets that we use in the paper. 

\paragraph{DUC} DUCs (2001 to 2007) \citep{nist2014duc} are modest MDS datasets in the news domain. Each year consists of 30-60 topics where each topic is composed of 10-30 source documents and 2-4 human-written (reference) summaries. Table~\ref{tab:duc_car} summarizes the dataset information for each separate year.

\paragraph{TAC} TAC (2008, 2009, 2010, 2011) \citep{nisttac} are similar to DUC datasets. Each year consists of approximately 45 topics, where each topic includes 10 source documents and 4 human-written summaries.

\paragraph{MultiNews~\citep{fabbri-etal-2019-multi}}
This dataset includes news articles and human-written summaries taken from the site \url{newser.com}. MultiNews is very large and the training, test and validation are composed of 44972, 5622 and 5622 topics respectively. The average number of source documents is 2.8 documents (range from 2 to 11 documents).

\paragraph{WCEP-10~\citep{gholipour-ghalandari-etal-2020-large}}
This dataset includes news events article retrieved from the Wikipedia Current Events Portal. The reference summary is human-written. The summary must be short (30-40 words) and each summary is one complete sentence. WCEP-10 is a truncated version of WCEP with a maximum number of source documents fix to 10. WCEP-10 is composed of 8158, 1022, 1020 topics for the train, validation and test set respectively.

\section{Additional Results}

\subsection{Assessment of Our Measure}
\label{app:human_comparison}

\begin{figure}
    \includegraphics[width=.48\textwidth]{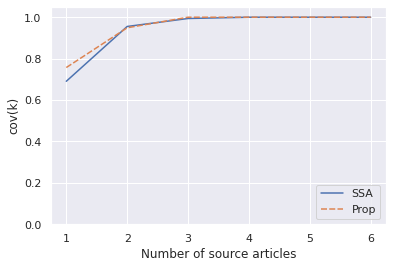} 
    \caption{Curves of $cov_k$ using the SSA annotations~\citep{ernst-etal-2021-summary} vs. our measure, for 9 topics from MultiNews}
    \label{fig:human-auto}
\end{figure}

Figure~\ref{fig:human-auto} shows the graphs of $cov_k$ according to human annotation in SSA and our measure. 

\begin{table}[t]
\centering
\begin{tabular}{lcccc}
\toprule
Year & \#Topics & \#Docs & \begin{tabular}[c]{@{}l@{}}\#Sums \\ by topic\end{tabular} & \#Sums \\ \midrule
2001 & 30 & 10 & 3 & 90  \\
2002 & 60 & 10 & 2 & 120 \\ 
2003 & 30 & 10 & 4 & 120 \\ 
2004 & 50 & 10 & 4 & 200 \\ 
2005 & 50 & 30 & 6 & 300 \\ 
2006 & 50 & 25 & 4 & 200 \\ 
2007 & 45 & 25 & 4 & 180 \\ \bottomrule
\end{tabular}
\caption{DUC's characteristics for each year.}
\label{tab:duc_car}
\end{table}

\begin{table}[t]
\centering
\begin{tabular}{lcccc}
\toprule
Year & \#Topics & \#Docs & \begin{tabular}[c]{@{}l@{}}\#Sums \\ by topic\end{tabular} & \#Sums \\ \midrule
2008 & 48 & 10 & 4 &  192 \\
2009 & 44 & 10 & 4 &  176 \\ 
2010 & 46 & 10 & 4 & 184 \\ 
2011 & 44 & 10 & 4 &  176\\ 
\bottomrule
\end{tabular}
\caption{TAC's characteristics for each year.}
\label{tab:TAC_car}
\end{table}

\subsection{Graphs of $cov_k$}
\label{app:graphs}
\begin{figure}[t]
\centering
    \includegraphics[width=.48\textwidth]{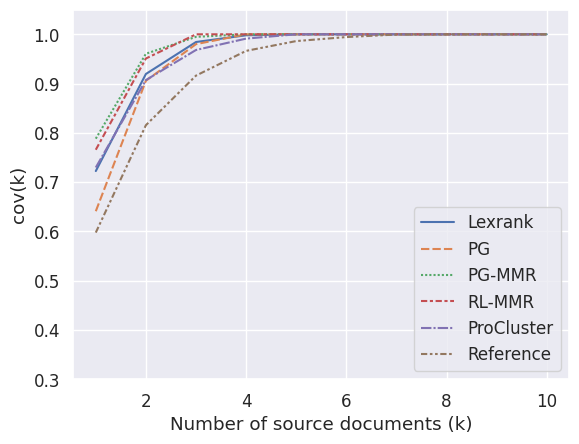}
    \caption{Curves of $cov_k$ on DUC 2004 systems summaries. The systems are trained on DUC 2003. The curves of ProCluster and PG are closer to the reference than Lexrank, PG-MMR and RL-MMR.}
    \label{fig:sys_duc2004}
\end{figure}

\begin{figure}[t]
    \centering
    \includegraphics[width=.48\textwidth]{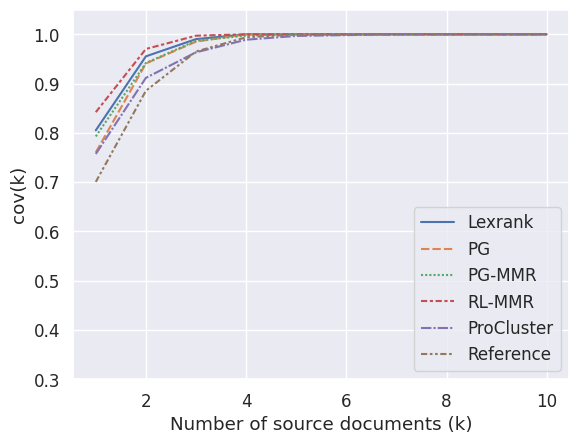}
    \caption{Curves of $cov_k$ on TAC 2011 systems summaries. The systems are trained on TAC 2008, 2009 and 2010. The curves of ProCluster and PG are closer to the reference than Lexrank, PG-MMR and RL-MMR. }
    \label{fig:tac}
\end{figure}

Figures~\ref{fig:sys_duc2004} and \ref{fig:tac} shows the curves of $cov_k$ on DUC 2004 and TAC 2011 systems summaries respectively. We can notice differents between the different system summaries, which indicate us that some models have higher degree of multi-text merging than other.

% \subsection{Model Performance vs. AAC}

% Figure~\ref{fig:ABC vs rouge} shows that the smaller the AAC the higher the ROUGE scores of the systems summary on MultiNews.

% \begin{figure}[t]
%     \centering

%     \includegraphics[width=.48\textwidth]{figures/AAC_vs_rouge.png}
%     \caption{ROUGE 2 and ROUGE L f scores of the PRIMERA system summaries on MultiNews test, according to AAC for each topic. The more the topic is dispersed, the smaller the ROUGE is.}
%     \label{fig:ABC vs rouge}
% \end{figure}

\section{Training on a Subset of Source Documents}
\label{app:mds_sds}

In order to properly assess the effect of fine-tuning an MDS model with only one or two source documents, we fine-tune also the MDS variant (given all source documents) with the same optimization steps. Table~\ref{tab:steps} presents the optimization and warm-up steps that were used for each dataset. 

% In some topics, multiple source documents might similarly cover

\begin{table}[t]
    \centering
    \begin{tabular}{lrr}
    \toprule
         Dataset & Total Steps & Warmup Steps \\
    \midrule
    DUC     & 20 & 2 \\
    TAC & 100 & 10 \\
    MultiNews & 10k & 1k \\
    WCEP-10 & 5k & 500 \\
    \bottomrule
    \end{tabular}
    \caption{Details of total and warm-up steps used for training the models with a single or all source documents, as described in Section~\ref{subsec:emperical}. We use the same number of steps for both experiments. }
    \label{tab:steps}
\end{table}

\end{document}